%% file: Formatting-Instructions-LaTeX-2025.tex
\documentclass[letterpaper]{article} % DO NOT CHANGE THIS
\usepackage{aaai25}  % DO NOT CHANGE THIS
\usepackage{times}  % DO NOT CHANGE THIS
\usepackage{helvet}  % DO NOT CHANGE THIS
\usepackage{courier}  % DO NOT CHANGE THIS
\usepackage[hyphens]{url}  % DO NOT CHANGE THIS
\usepackage{graphicx} % DO NOT CHANGE THIS
\usepackage{natbib}  % DO NOT CHANGE THIS AND DO NOT ADD ANY OPTIONS TO IT
\usepackage{caption} % DO NOT CHANGE THIS AND DO NOT ADD ANY OPTIONS TO IT
\usepackage{algorithm}
\usepackage{algorithmic}

\usepackage{booktabs} 
\usepackage{tabularx}
\usepackage{multirow}
\usepackage{makecell}
\usepackage{colortbl}
\usepackage{array}
\usepackage{subcaption}
\usepackage{amsmath}
\usepackage{amssymb}
\usepackage[table,xcdraw]{xcolor}
\usepackage{pifont}

\usepackage{newfloat}
\usepackage{listings}
\DeclareCaptionStyle{ruled}{labelfont=normalfont,labelsep=colon,strut=off} % DO NOT CHANGE THIS
\floatstyle{ruled}
\newfloat{listing}{tb}{lst}{}
\floatname{listing}{Listing}
\title{Memory Efficient Matting with Adaptive Token Routing}
\author{
    \author{Yiheng Lin\textsuperscript{\rm 1, \rm 2},
    Yihan Hu\textsuperscript{\rm 1, \rm 2, \rm 4},
    Chenyi Zhang\textsuperscript{\rm 1, \rm 2, \rm 4},\\
    Ting Liu\textsuperscript{\rm 4},
    Xiaochao Qu\textsuperscript{\rm 4},
    Luoqi Liu\textsuperscript{\rm 4},
    Yao Zhao\textsuperscript{\rm 1, \rm 2, \rm 3}\thanks{Corresponding author.},
    Yunchao Wei\textsuperscript{\rm 1, \rm 2, \rm 3}}
}
\affiliations{
    \textsuperscript{\rm 1}Institute of Information Science, Beijing Jiaotong University,\\
    \textsuperscript{\rm 2}Visual Intelligence + X International Joint Laboratory of the Ministry of Education,\\
    \textsuperscript{\rm 3}Pengcheng Laboratory, Shenzhen, China,\\
    \textsuperscript{\rm 4}MT Lab, Meitu Inc.\\
    \{yiheng.lin, yzhao, yunchao.wei\}@bjtu.edu.cn
}

\usepackage{bibentry}
\begin{document}

\maketitle

\begin{abstract}
Transformer-based models have recently achieved outstanding performance in image matting. However, their application to high-resolution images remains challenging due to the quadratic complexity of global self-attention. To address this issue, we propose MEMatte, a \textbf{m}emory-\textbf{e}fficient \textbf{m}atting framework for processing high-resolution images. MEMatte incorporates a router before each global attention block, directing informative tokens to the global attention while routing other tokens to a Lightweight Token Refinement Module (LTRM). Specifically, the router employs a local-global strategy to predict the routing probability of each token, and the LTRM utilizes efficient modules to simulate global attention. Additionally, we introduce a Batch-constrained Adaptive Token Routing (BATR) mechanism, which allows each router to dynamically route tokens based on image content and the stages of attention block in the network. Furthermore, we construct an ultra high-resolution image matting dataset, UHR-395, comprising 35,500 training images and 1,000 test images, with an average resolution of $4872\times6017$. This dataset is created by compositing 395 different alpha mattes across 11 categories onto various backgrounds, all with high-quality manual annotation. Extensive experiments demonstrate that MEMatte outperforms existing methods on both high-resolution and real-world datasets, significantly reducing memory usage by approximately 88\% and latency by 50\% on the Composition-1K benchmark. Our code is available at \url{https://github.com/linyiheng123/MEMatte}.

\end{abstract}



\section{Introduction}

Natural image matting is a crucial task in computer vision that aims to accurately separate the foreground object by predicting the alpha matte. Given a foreground $F\in\mathbb{R}^{H\times W\times C}$, a background $B\in\mathbb{R}^{H\times W\times C}$, and an alpha matte $\alpha\in\mathbb{R}^{H\times W}$, a natural image $I\in \mathbb{R}^{H\times W\times C}$ can be represented as:
\begin{equation}
\label{eq:mixing}
I_{i} = \alpha_{i}F_{i} + (1-\alpha_i)B_{i}, \quad \alpha\in \left [ 0,1 \right],
\end{equation}
where $H$, $W$, and $C$ denote the height, width, and channel of the image, respectively, and $i$ denotes the pixel index.

In recent years, vision transformers (ViT) have achieved impressive results in image matting \cite{hu2025diffusion, yao2024vitmatte} due to their long-range modeling capability. Nevertheless, applying ViT-based methods on memory-limited GPUs is challenging as global attention has quadratic complexity and full-resolution input for matting is crucial. As shown in Figure \ref{fig:downsample}, image matting demands high precision for each pixel, and simply downsampling the image leads to distortion. Moreover, a patch-based crop-and-stitch manner is also inappropriate as it destroys global context and introduces artifacts. These limitations raise the question: how can we reduce the memory cost of global attention while maintaining full-resolution input?

\input{Figures/downsample}

A common approach to solve this problem involves token pruning \cite{rao2021dynamicvit, liang2022evit} and token merging \cite{bolya2022tome}, where a fixed ratio of redundant tokens is progressively discarded at specific network stages. However, this paradigm presents two problems: \textbf{P1)} Applying a fixed ratio results in poor performance and inefficiency. For instance, a fixed ratio causes unnecessary computational overhead for simple images and performance degradation for complex images. Additionally, using a fixed ratio across different network stages is inefficient, since global attention behaves like convolution in the early stages \cite{park2022vision, ghiasi2022vision}. \textbf{P2)} Progressive token discarding degrades performance, as discarded tokens are excluded from subsequent calculations. This approach risks losing important tokens in the middle layers, therefore making it unsuitable for precise predictions.

To address these issues, we propose MEMatte. The core idea of MEMatte is to \textbf{route informative tokens to the global attention branch while directing other tokens to an efficient branch}. Specifically, a router utilizing a local-global strategy is inserted before each global attention to predict the routing probability for each token. Then a Batch-constrained Adaptive Token Routing mechanism (BATR) is introduced to make the routing decision based on the predicted probability. BATR adaptively adjusts the number of tokens routed to the two branches according to image complexity and the network stage, thus addressing the problem \textbf{P1}. Notably, BATR reselects the branch for all tokens at each block, which solves the problem \textbf{P2}. We also design a lightweight token refinement module (LTRM) as the efficient branch, which employs depth-wise convolution and efficient channel attention to update the uninformative tokens.

\input{Figures/1k_8k.tex}
By allocating different computational resources to different tokens, MEMatte significantly decreases memory usage during inference. As shown in Figure \ref{fig:1k_8k}, MEMatte can process over 4K resolution images on commonly used consumer-level GPUs, such as the RTX 1060, and 8K resolution images on the RTX 3090 GPU. The memory consumption of MEMatte is significantly lower than that of the ViT-based method ViTMatte \cite{yao2024vitmatte}, and even more efficient than the Swin-based method MatteFormer \cite{park2022matteformer} and the CNN-based methods GCA \cite{li2020natural} and FBA \cite{forte2020f}.

We also evaluate MEMatte on the widely-used benchmark Composition-1K, where it reduces memory usage by approximately 88\% and inference latency by 50\% compared to ViTMatte. Moreover, MEMatte achieves state-of-the-art results on both high-resolution and real-world datasets.

Our contributions are summarized as follows:
\begin{itemize}
    \item We propose MEMatte, which integrates a router, a Batch-constrained Adaptive Token Routing mechanism, and a lightweight token refinement module, effectively reducing computational overhead during inference.
    
    \item We conduct extensive experiments on widely used benchmarks. MEMatte achieves state-of-the-art performance on both high-resolution and real-world datasets.

    \item We contribute UHR-395, an ultra high-resolution image matting dataset with manual annotations. UHR-395 includes 395 different foreground objects across 11 categories, with an average resolution of 4872$\times$6017. 
    % To the best of our knowledge, this is the highest-resolution publicly available matting dataset.

\end{itemize}

\section{Related Works}
\subsection{Natural Image Matting}
Traditional image matting methods focused on sampling-based \cite{gastal2010shared, he2011global} and propagation-based approaches \cite{chen2013knn, he2010fast, levin2007closed} to estimate alpha values using known foreground and background colors. With the rise of deep learning, CNN-based methods \cite{xu2017deep, hao2019indexnet, li2020natural, forte2020f, yu2021mask, sun2021semantic, li2024disentangled} have significantly advanced matting performance. Recently, inspired by the success of transformers \cite{vaswani2017attention, dosovitskiy2020image, liu2021swin, li2022exploring, zhang2023controlvideo, liu2024forgery} in NLP and vision tasks, image matting methods \cite{park2022matteformer, cai2022transmatting, dai2022boosting, yao2024vitmatte, li2024vmformer} have incorporated transformers, yielding impressive performance.

\subsection{High Resolution Image Matting}
Handling high-resolution images in matting is challenging. HDMatte \cite{yu2021high} introduced a Cross-Patch Contextual module to model contextual dependencies. BGMv2 \cite{lin2021real} used a small patch replacement strategy to process high-resolution images. Both of these patch-based methods rely solely on convolutional neural networks, which are ineffective within transformer models because they disrupt the global receptive field of global attention. SparseMat \cite{sun2023ultrahigh} utilized dilation and erosion operations to identify active pixels and refined them with sparse convolution \cite{engelcke2017vote3deep}. However, this approach often fails to accurately identify pixels requiring correction, particularly with transparent objects. 
% Despite these advances, ViT-based methods for high-resolution matting have yet to be explored.

\subsection{Token Compression}
Common token compression methods typically reduce the number of tokens by token pruning or merging. DynamicViT \cite{rao2021dynamicvit} maintained a score vector to prune redundant tokens hierarchically. EViT \cite{liang2022evit} proposed a token reorganization method to identify and fuse image tokens. ToMe \cite{ bolya2022tome} introduced a matching algorithm for merging similar tokens. However, these methods encounter out-of-memory errors in high-resolution matting as they retain all tokens during the early stages of the network.

\input{Figures/framework}

\section{Methodology}

\subsection{Preliminaries}
In this section, we review the mechanism of global attention. Let the input image be $I\in\mathbb{R}^{H\times W \times C}$. The image is first encoded into tokens $X \in \mathbb{R}^{N \times D}$, where $N = HW/{p^2}$ represents the number of tokens and $p$ represents the token size. In global attention, $X$ is projected into query $Q \in \mathbb{R}^{N\times D_{q}}$ , key $K \in \mathbb{R}^{N\times D_{k}}$, and value $V \in \mathbb{R}^{N\times D_{v}}$. The operation of global self-attention is defined as:
\begin{equation}
\label{attention}
Attention(Q, K, V) = Softmax(\frac{QK^{T}}{\sqrt{d}})V,
\end{equation}
where the similarity score computed between query-key pairs leads to an $O(N^2)$ complexity. Specifically, the memory required to store the attention map is approximately $4hN^2$ bytes, where $h$ denotes the number of heads. 
% For an 8K input image, this requires around 400GB of memory. 

These computational and memory challenges underscore the necessity of a more efficient mechanism, prompting us to design MEMatte, which addresses this challenge through adaptive token routing.

\subsection{Adaptive Token Routing}
The overall framework of MEMatte is illustrated in Figure \ref{fig:framework}. In MEMatte, each transformer block consists of four main components: the \textbf{Router} module, the Batch-constrained Adaptive Token Routing (\textbf{BATR}) mechanism, the global attention branch, and the Lightweight Token Refinement Module (\textbf{LTRM}) branch. After the image tokens enter the transformer block, the router module predicts the routing probability for each token. The BATR then makes the final routing decision, directing tokens to one of the two branches. Finally, the tokens from both branches are merged together.

\noindent\textbf{Router.}
The router module utilizes a local-global strategy to calculate the routing probability for each token. Given image tokens $X\in \mathbb{R}^{N\times D}$, let $x_i\in \mathbb{R}^{D}$ represent the $i$-th token. First, $x_i$ is projected through a layer normalization $LN$ and a linear layer $f_{\theta}$:
\begin{equation}
z_i = f_{\theta}(LN(x_i)) \in \mathbb{R}^{D}.
\end{equation}
Next, $z_i$ is split along the channel dimension into a local feature $z_{i}^{l}$ and a global feature $z_{i}^{g}$. The local feature $z_{i}^{l}$ retains the first half of the channels:
\begin{equation}
z_{i}^{l} = z_{i}[\ :D/2\ ] \in \mathbb{R}^{D/2},
\end{equation}
which encodes the information of a single token. The global feature $z_i^g$ is obtained by averaging the second half of the channels over all tokens:
\begin{equation}
z_i^g = \frac{1}{N}\sum_{i=1}^{N}z_i[\ D/2:\ ] \in \mathbb{R}^{D/2},
\end{equation}
which encodes the global contextual information. Then $z_i^l$ and $z_i^g$ are concatenated along the channel dimension: 
\begin{equation}
    z_i^{\prime} = Cat(z_i^l, z_i^g)\in \mathbb{R}^{D}.
\end{equation}
This local-global strategy ensures that $z_i^{\prime}$ contains both local and global contextual information. Finally, $z_i^{\prime}$ is passed to a linear layer $f_{\theta}^{\prime}$ and a LogSoftmax layer $LS$ to predict the routing probability $p_i$ for the $i$-th token:
\begin{equation}
    log\ p_i = LS(f_{\theta}^{\prime}(z_i^{\prime}))\in \mathbb{R}^{2}.
\end{equation}
where $p_{i,0}$ and $p_{i,1}$ represent the probabilities of routing the $i$-th token to the LTRM branch and the global attention branch, respectively.

\noindent\textbf{Batch-constrained Adaptive Token Routing.} 
BATR possesses two key features that address the issues outlined in the introduction. First, BATR reselects the branch for all tokens in each block, preventing the loss of informative tokens. Second, instead of setting a fixed ratio, BATR adaptively adjusts the number of tokens routed to the two branches based on image complexity and the network stage. Below, we provide a detailed explanation of how BATR works.

Consider a mini-batch of $B$ samples and $M$ stages of the model, $p_i^{b,m}\in \mathbb{R}^2$ represents the routing probability of the $i$-th token, where $b$ denotes the sample index and $m$ denotes the stage index. The routing decision $\delta_i^{b,m}\in \{0,1\}$ is determined as follows:
\begin{equation}
    \delta_i^{b,m} = \left\{
    \begin{array}{ll}
    % \begin{matrix}
    G(p_i^{b,m}) & \mathrm{Training}, \\ [4pt]
    A(p_i^{b,m}) & \mathrm{Inference}.
    % \end{matrix}
    \end{array}
    \right.
\end{equation}
Here, $G$ denotes the Gumbel-Softmax function, which ensures that the routing process is differentiable. $A$ represents the Argmax function, which avoids setting a fixed routing ratio. A value of $\delta_i^{b,m}=0$ indicates that the $i$-th token is routed to the LTRM, while $\delta_i^{b,m}=1$ indicates that the token is routed to global attention.

To ensure that the token routing process is content-aware and stage-aware, we first calculate the average ratio $\gamma$ of tokens routed to attention across all routers at the batch level:
\begin{equation}
    \gamma = \frac{1}{BMN}\sum_{b=1}^{B}\sum_{m=1}^{M}\sum_{i=1}^{N}\delta_i^{b,m} \in [0,1].
\end{equation}
During training, by constraining the value of $\gamma$ and incorporating matting loss, the model learns to route fewer tokens to the attention branch at the appropriate samples and stages.

\noindent\textbf{Lightweight Token Refinement Module.}
Due to the high accuracy demands of image matting, simply skipping the global attention for uninformative tokens can result in a performance decline. To address this issue, we employ a Lightweight Token Refinement Module (LTRM), inspired by the Skip-At method \cite{venkataramanan2023skip}, to effectively process these tokens.

Given the representation of tokens $X\in \mathbb{R}^{N\times D}$, we first project it using a linear layer $f_{\theta_1}$. Next, we apply a depth-wise convolution $DWC$ to enhance the extraction of texture information in a lightweight manner. Finally, the output is further refined by a linear layer $f_{\theta_2}$ and an efficient channel attention module $ECA$ \cite{wang2020eca}. The updated tokens $Z$ can be formalized as:
\begin{equation}
    Z = ECA(f_{\theta_{2}}(DWC(f_{\theta_{1}}(X)))) \in \mathbb{R}^{N\times D}.
\end{equation}

\noindent\textbf{Adaptation to Ultra High-Resolution Images.}
For ultra high-resolution images, the routers may direct an excessive number of tokens to the attention branch as they are content-aware and stage-aware. To mitigate this, we set a maximum number of tokens $k$ during inference. If the number of tokens routed to attention exceeds $k$, only the top $k$ most important tokens are retained based on the scores $p_{i,1}$ provided by the router. Our ablation study shows that this approach has little impact on performance, as the remaining tokens are still updated by the LTRM.

\subsection{Training Objectives}

\noindent\textbf{Distillation loss.}
To enable the LTRM to emulate the global attention mechanism, we utilize the ViT from ViTMatte as the teacher model. The teacher model guides the output of the MEMatte's ViT to closely approximate that of the original ViT. Given the output features $F_t\in \mathbb{R}^{N\times D}$ from teacher ViT and $F_s\in \mathbb{R}^{N\times D}$ from MEMatte ViT, the distillation loss is defined as:
\begin{equation}
    \mathcal{L}_{distill} = \frac{1}{N} \sum_{i=1}^{N} (F_{t,i} - F_{s,i})^2,
\end{equation}
where i denotes the token index.

\noindent\textbf{Compression loss.}
To reduce the number of tokens routed to the global attention branch, we introduce a target compression degree $\rho\in \left [0,1  \right ]$ to constrain the value of $\gamma$. Here, $\rho$ is a predefined hyperparameter. The loss item can be formulated as:
\begin{equation}
	\mathcal{L}_{compress} = (\rho - \gamma)^2.
\end{equation}
A smaller value of $\rho$ results in fewer tokens being routed to the global attention branch.

\noindent\textbf{Total loss.}
We combine the aforementioned losses with the matting loss from ViTMatte to formulate the total loss:
\begin{equation}
    \mathcal{L}_{total} = \mathcal{L}_{matting} + \mathcal{L}_{distill} + \mathcal{L}_{compress}.
\end{equation}

\subsection{Ultra High-Resolution Dataset}
As illustrated in Table \ref{tab:dataset}, we select several widely used image matting datasets for comparison, including DIM \cite{xu2017deep}, Distinctions-646 \cite{qiao2020attention}, AIM-500 \cite{li2021deep}, PPM-100 \cite{ke2022modnet}, and Transparent-460 \cite{cai2022transmatting}. These datasets are not suitable benchmarks for high-resolution matting as they fail to satisfy the requirements for both category diversity and high resolution. While Transparent460 and PPM100 offer higher resolutions, the former is primarily composed of transparent objects, and the latter focuses on portraits. Conversely, DIM, Distinctions-646, and AIM-500 suffer from insufficient resolution.

To address these limitations, we propose UHR-395, an ultra high-resolution dataset. UHR-395 comprises 395 foreground objects across 11 categories, such as animals with fur, fire, humans, plants, glass, water, etc. Each object is meticulously annotated by a professional team, with the annotations undergoing multiple validation rounds to ensure high quality. We divide these objects into 355 for training and 40 for testing, producing 35,500 training images and 1,000 test images following the rules in DIM. More details are available in the supplementary materials.

\input{Tables/dataset}

\input{Tables/synthetic.tex}

\section{Experiments}

\subsection{Datasets and Evaluation}
\textbf{Synthetic Dataset.} Due to the high annotation costs, available datasets for image matting are limited. DIM and Distinction-646 composite foreground images with background images from the COCO \cite{lin2014microsoft} and VOC2012 \cite{everingham2010pascal} datasets using Eq. ~\ref{eq:mixing} for both training and evaluation. Following other matting methods, we use Composition-1K to represent the DIM test set.

\noindent\textbf{High-Resolution Dataset.} Due to the large resolution of our UHR-395 test set (average 5318$\times$7051), most popular image matting methods encounter out-of-memory errors. For a more detailed comparison, we select 344 images larger than 4K resolution from the PPM-100 and Transparent-460 datasets, naming this subset PPT-344, with an average resolution of $3962 \times 4170$.

\noindent\textbf{Real-world Dataset.} AIM500 is a real-world test dataset that encompasses various objects, including portraits, animals, transparent objects, and fruits. We choose this dataset to evaluate the ability of existing image matting methods to generalize to real-world scenarios.

Following existing methods, we train our model on the DIM training dataset and evaluate it on synthetic, high-resolution, and real-world datasets to demonstrate the superiority of our method in terms of low memory cost and strong generalization ability. We compare our method with popular matting methods introduced in related works, using four commonly used metrics: Sum of Absolute Differences (SAD), Mean Square Error (MSE), Gradient loss (Grad), and Connectivity loss (Conn). Note that lower values of these metrics indicate better alpha matte quality. All experiments are performed on the RTX 3090. More implementation details are in the supplementary material.

\subsection{Main Results}
\noindent\textbf{Results on Synthetic Dataset.}
The quantitative results on the synthetic dataset are shown in Table \ref{tab:synthetic}. Our method significantly reduces the computational overhead of the ViTMatte baseline, the average memory usage decreased by 5.49 GB (\textbf{-88.5\%}) when using ViT-S and by 11.04 GB (\textbf{-88.1\%}) when using ViT-B. Despite this significant memory cost reduction, the performance degradation on the Composition-1K test set is only around $5\%$. Moreover, the performance exhibits an improvement when generalized to the Distinction-646 test set.
\input{Figures/uhrim.tex}

\input{Tables/UHRIM.tex}

\noindent\textbf{Results on High-Resolution Dataset.} Due to the high resolution of our proposed UHR-395 test set, we evaluate several popular matting methods under two different settings to avoid out-of-memory errors. In the first setting, inference is performed on downsampled images ($i.e.$ 1024$\times$1024) and then upsampled to the original resolution. In the second setting, a crop-and-stitch strategy is used, conducting inference on cropped patches ($i.e.$ 512$\times$512). The quantitative results on our UHR-395 test set, shown in Table \ref{tab:uhrim}, clearly demonstrate the superiority of our method. Besides, as visualized in Figure \ref{fig:uhrim}, the downsampling strategy leads to distortion, while the crop-and-stitch strategy introduces artifacts. In contrast, our method produces high-quality output.

Additionally, we evaluate these methods on the PPT-344 dataset, which has a lower resolution, enabling a comprehensive comparison with full-resolution input. ViTMatte still employs the two settings above due to out-of-memory error. As presented in Table \ref{tab:PPT344}, our method consistently outperforms the others.
\input{Tables/PPT344.tex}

\noindent\textbf{Results on Real-world Dataset.} Given the lack of real-world images for training, existing methods typically train their models on synthetic images. Consequently, the performance of the model on real-world dataset becomes critically important. As shown in Table \ref{tab:aim500}, our method demonstrates great generalization ability on real-world images and achieves state-of-the-art performance. We also provide visualizations in supplementary materials.

\input{Tables/aim500.tex}

\noindent\textbf{Visualization of Token Routing}.
To further investigate the routing mechanism of MEMatte, we visualize the tokens routed to global attention by different routers in  Figure \ref{fig:sparsity}. We observe several key points: (1) Routers preferentially route informative tokens to global attention, such as the edges of target objects and areas with complex textures. (2) Across different scenarios, the distribution of selected tokens remains sparse, yet the results are consistently excellent. (3) The first router directs only a few tokens to global attention. This is reasonable because global attention in the early network stages behaves like convolution \cite{park2022vision, ghiasi2022vision}, preferring to capture general edges and textures, which are effectively handled by the LTRM. 

\input{Figures/sparsity.tex}

\subsection{Ablation study}
\label{sec:ablation}

\noindent\textbf{Comparisons of Different Token Compression Methods.} To assess the efficacy of our adaptive token routing method in comparison to other token compression methods, we replace the backbone of the ViTMatte with efficient ViT proposed by popular token pruning or merging methods, such as DynamicViT, EViT, and ToMe. As shown in Table \ref{tab:routing_mechanism}, these methods exhibit poor performance as they progressively discard a fixed ratio of redundant tokens at specific network stages, leading to the issues mentioned in the introduction. Additionally, these methods also exhibit high memory usage as they retain most tokens during the early stages.

\input{Tables/routing_mechanism}

\noindent\textbf{Effect of Reducing Maximum Token Number.} Intuitively, reducing maximum token number $k$ might lead to a significant performance drop. However, as shown in Figure \ref{fig:reduce_token_number_bar},  as $k$ decreases, the performance metrics remain stable while the memory cost drops rapidly. We attribute this robustness to the effectiveness of the token scoring approach and LTRM.

\input{Figures/reduce_token_number_bar}

\noindent\textbf{Effect of Target Compression degree.}
Table \ref{tab:target_compression_degree} shows the effects of target compression degree $\rho$. As the $\rho$ decreases from 0.75 to 0.1, performance declines until reaching 0.25, beyond which it deteriorates quickly. Consequently, we select 0.25 as the default setting during training for a better trade-off between performance and efficiency.

\input{Tables/target_compression_degree}

\noindent\textbf{Effect of Distillation and LTRM.} Table \ref{tab:components} reports the effectiveness of the distillation loss and LTRM. Using either the distillation loss or LTRM alone yields similar performance to using neither of them. However, combining both of them results in a significant performance improvement. This suggests that the teacher model successfully teaches the LTRM to imitate the mechanism of global attention.

\input{Tables/components.tex}

\section{Conclusion}
In this paper, we propose MEMatte and an ultra high-resolution dataset UHR-395. MEMatte significantly reduces memory usage and latency during inference compared to existing matting methods, achieving state-of-the-art performance on both high-resolution and real-world datasets. These results highlight the robustness of our token routing mechanism and LTRM. We hope that MEMatte will promote further research on high-resolution matting.

\section{Acknowledgments}
This work was supported by the National Natural Science Foundation of China (No.92470203, No.U23A20314), Beijing Natural Science Foundation (No. L242022), the Fundamental Research Funds for the Central Universities (2024XKRC082).

\bibliography{aaai25}

\end{document}

%% file: Figures/downsample.tex
\begin{figure}
    \centering
    \includegraphics[width=0.46\textwidth]{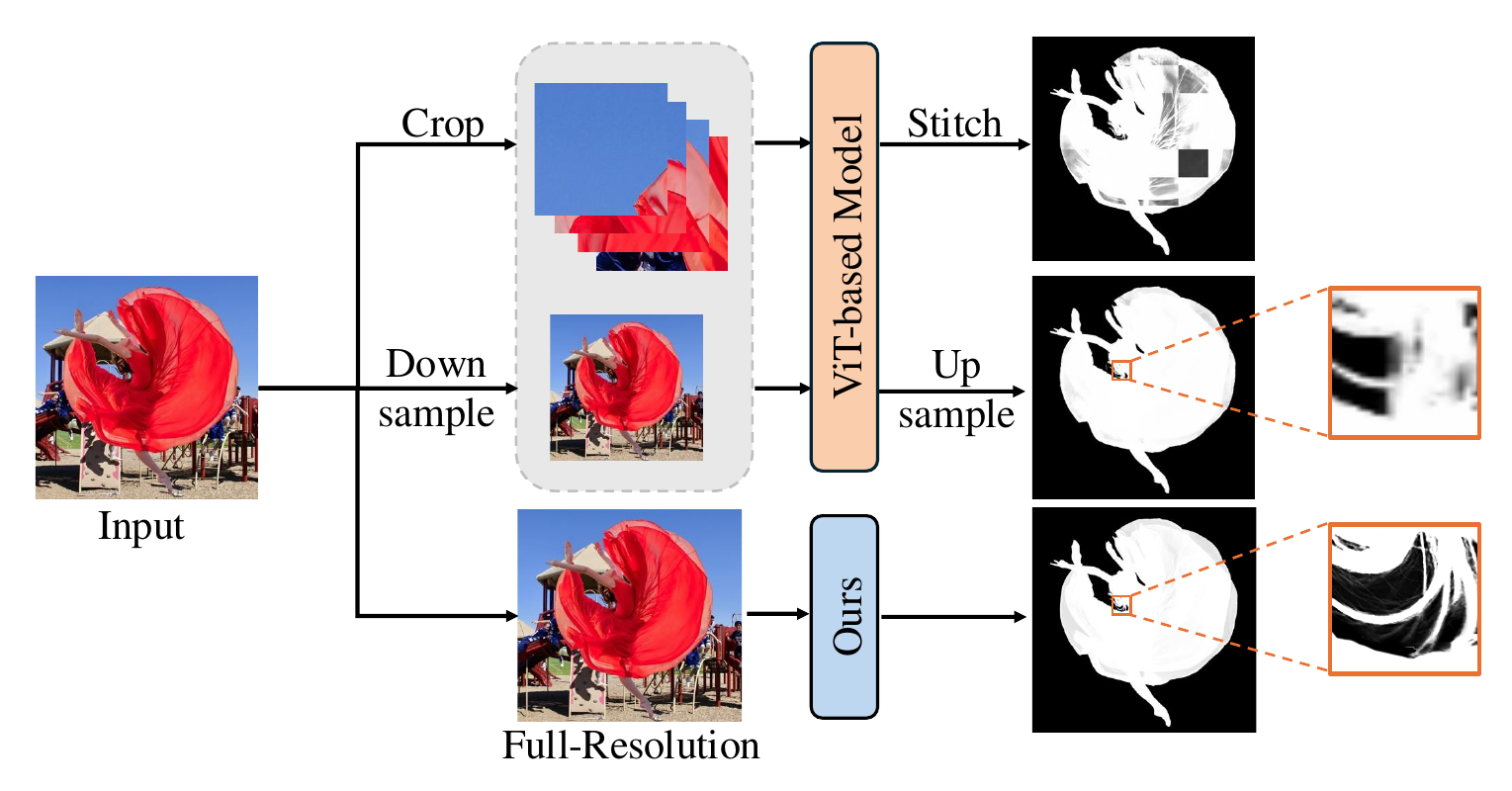}
    \caption{Illustration of the importance of full-resolution input in image matting. The crop-and-stitch manner introduces artifacts, while downsampling manner causes distortion.
    % Visualization of the predictions of original resolution images and downsampled images entering the model.
    }
    \label{fig:downsample}
\end{figure}

%% file: Figures/1k_8k.tex
\begin{figure}
    \centering
    \captionsetup{aboveskip=10pt, belowskip=2pt}
    \includegraphics[width=0.47\textwidth]{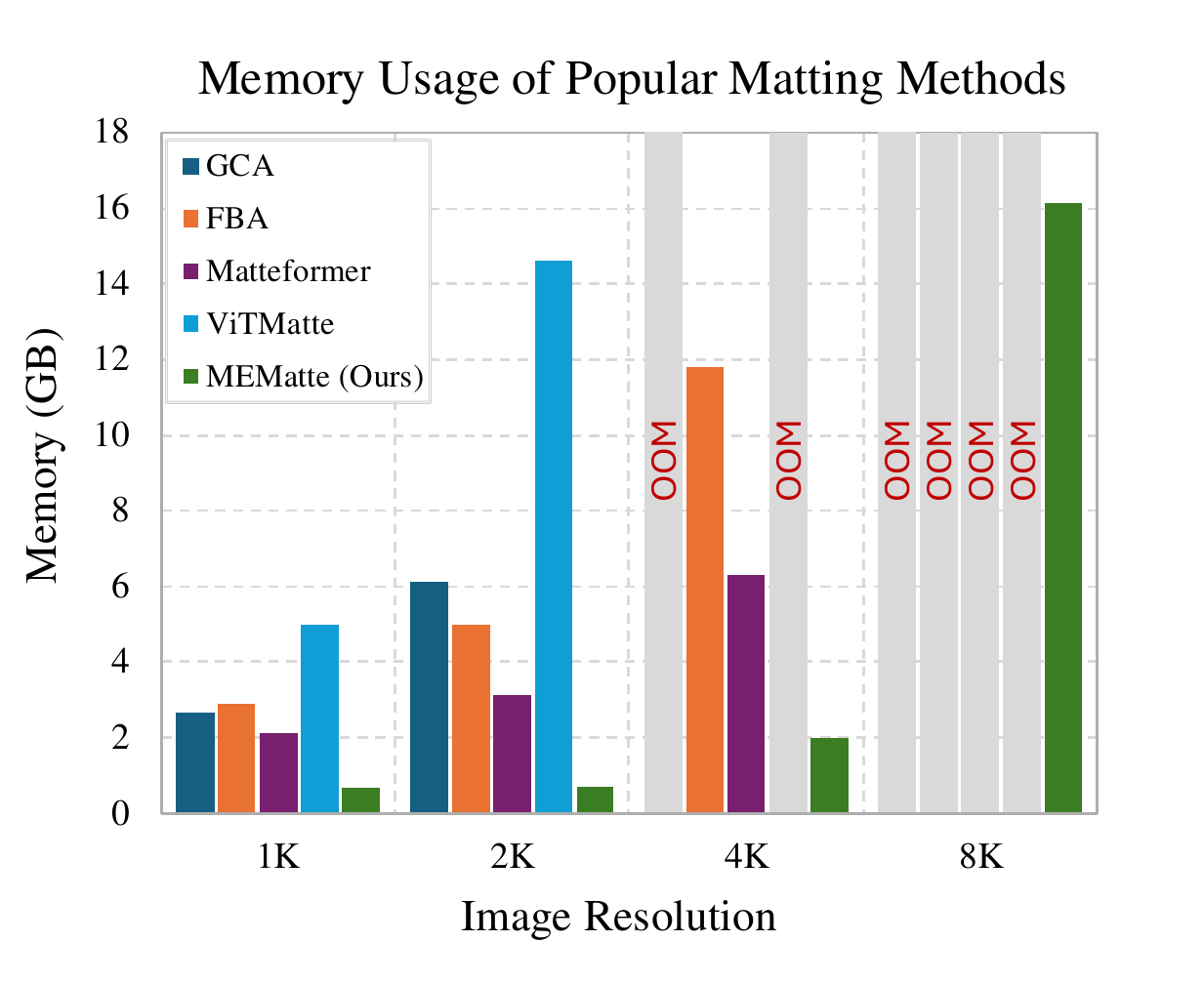}
    \caption{Memory Usage / Image Resolution. OOM denotes an out-of-memory error encountered on an RTX 3090 GPU. The huge increase in memory usage from 4K to 8K of MEMatte is due to the 4x tokens and the quadratic complexity of attention mechanisms. Despite this, MEMatte is capable of processing 8K images on the RTX 3090.}
    \label{fig:1k_8k}
\end{figure}

%% file: Figures/framework.tex
\begin{figure*}[htb]
    \centering
    \includegraphics[width=0.98\textwidth]{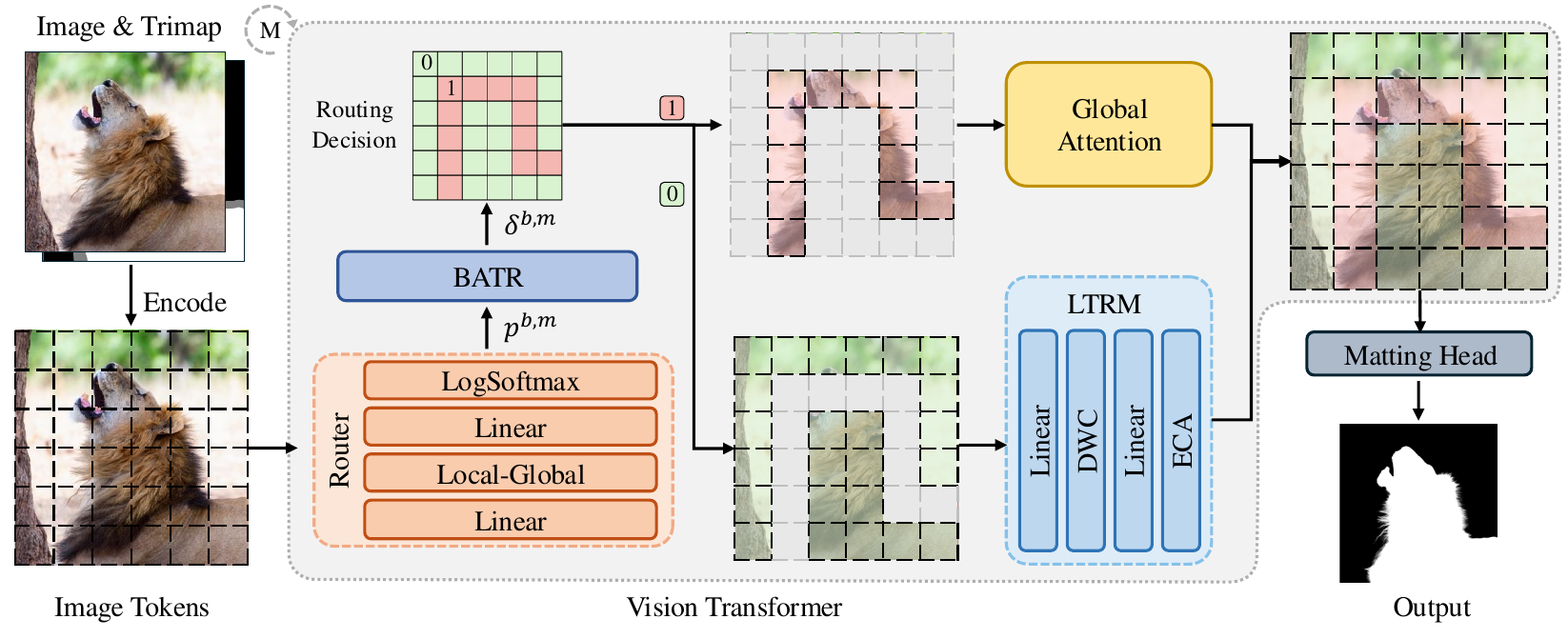}
    \caption{Overall framework of the proposed MEMatte. The router module is inserted before global attention to predict the routing probability $p^{b,m}$ for each token. The BATR then makes the routing decision $\delta^{b,m}$ based on $p^{b,m}$.}
    \label{fig:framework}
\end{figure*}

%% file: Tables/dataset.tex
\begin{table}[htb]
\centering
\begin{tabular}{l|ccc}
\toprule
Datasets               & Avg Resolution         & FN & MC   \\ \midrule
DIM                    & 1082$\times$1297       & 481 & \ding{51}     \\
Distinctions-646       & 1727$\times$1565       & 646 & \ding{51}     \\
AIM-500                & 1260$\times$1397       & 500 & \ding{51}     \\
PPM-100                & 2875$\times$2997       & 100 & \ding{55}     \\
Transparent-460        & 3772$\times$3804       & 460 & \ding{55}    \\
\rowcolor[HTML]{d7dbdd} 
UHR-395 (Ours)           & \textbf{4872$\times$6017}       & 395 &  \ding{51}    \\ \bottomrule

\end{tabular}
\caption{Comparison between different public matting datasets. FN denotes the number of foreground and MC indicates multiple categories. Our dataset is marked in gray.}
\label{tab:dataset}
\end{table}

%% file: Tables/synthetic.tex
\begin{table*}[ht]
\centering
\setlength{\tabcolsep}{3.9pt} 
\begin{tabular}{l|cccc|cccc|lll}
\toprule
Datasets & \multicolumn{4}{c|}{Composition-1K} & \multicolumn{4}{c|}{Distinctions-646}  & \multirow{2}{*}{Mem$_{/\text{GB}}$} & \multirow{2}{*}{Latency$_{/\text{ms}}$} & \multirow{2}{*}{GFLOPs}  \\ 
Methods      & SAD $\downarrow$ & MSE $\downarrow$ & Grad $\downarrow$ & Conn $\downarrow$ & SAD $\downarrow$ & MSE $\downarrow$ & Grad $\downarrow$ & Conn $\downarrow$ &  & &  \\ \midrule 
DIM           & 50.4   & 14.0   & 31.0    & 50.8       & 58.73 & 14.77 & 77.32 & 59.89 & 6.05 & 142.00 & 1591.52 \\
IndexNet      & 45.8   & 13.0   & 25.9    & 43.7       & 46.73 & 9.63 & 43.34 & 46.86 & 2.71 & 94.14 & 255.29 \\  
GCA           & 35.28  & 9.00   & 16.90   & 32.50     & 35.33 & 18.4 & 28.78 & 34.29 & 3.01 & 179.251 & 939.45 \\
HDMatt        & 33.5   &7.3     & 14.5    &29.9        & - & - & - & - & - & - & -\\
MGMatting     & 31.5   & 6.8    & 13.5    & 27.3       & 33.24 & 4.5 & 20.31 & 25.49 & 1.96 & 112.01 & 434.92 \\
SIM           & 28.0   & 5.8    & 10.8    & 24.8       & 23.60 & 3.93 & 20.05 & 22.20 & 3.49 & 550.14 & 2200.43 \\
FBA           & 25.8   & 5.2    & 10.6    & 20.8       & 32.37 & 5.47 & 24.15 & 31.70 & 3.08 & 538.03 & 1497.57 \\
TransMatting  & 24.96  & 4.58   & 9.72    & 20.16     & - & - & - & - & - & - & - \\
Matteformer   & 23.80  & 4.03   & 8.68    & 18.90     & 23.90 & 8.16 & 12.65 & 18.90 & 2.20 & 220.83 & 546.61 \\
RMat          & 22.87  & 3.9    & 7.74    & 17.84     & - & - & - & - & - & - & - \\ \midrule
ViTMatte-S    & 21.46  & 3.3    & 7.24    & 16.21     & 23.43 & 2.41 & 11.49 & 19.02 & 6.20 & 186.00 & 743.03 \\
\rowcolor[HTML]{d7dbdd} 
MEMatte-S        & 21.90  & 3.37   & 7.43    & 16.77    & 21.54 & 2.46 & 11.06 & 18.27 & \textbf{0.71}$_{\mathbf{88.5\%\downarrow}}$ & \textbf{84.99}$_{\mathbf{54.3\%\downarrow}}$ & 661.14 \\ \midrule
ViTMatte-B    & 20.33  & 3.0    & 6.74    & 14.78     & 22.09 & 1.93 & 9.59 & 17.26 & 12.53 & 340.15 & 1710.14\\
\rowcolor[HTML]{d7dbdd} 
MEMatte-B        & 21.06  & 3.11   & 6.70    & 15.71    & 20.02 & 1.92 & 9.39 & 16.97 & \textbf{1.49}$_{\mathbf{88.1\%\downarrow}}$ & \textbf{178.90}$_{\mathbf{47.4\%\downarrow}}$ & 1432.92 \\ \bottomrule

\end{tabular}
\caption{Quantitative results on Synthetic Dataset. The memory usage, latency, and GFLOPs are evaluated using the Composition-1K dataset, averaged across images. A dash (-) indicates no official code release. Our method is marked in gray.}
\label{tab:synthetic}
\end{table*}

%% file: Figures/uhrim.tex
\begin{figure*}
    \centering
    \includegraphics[width=0.98\textwidth]{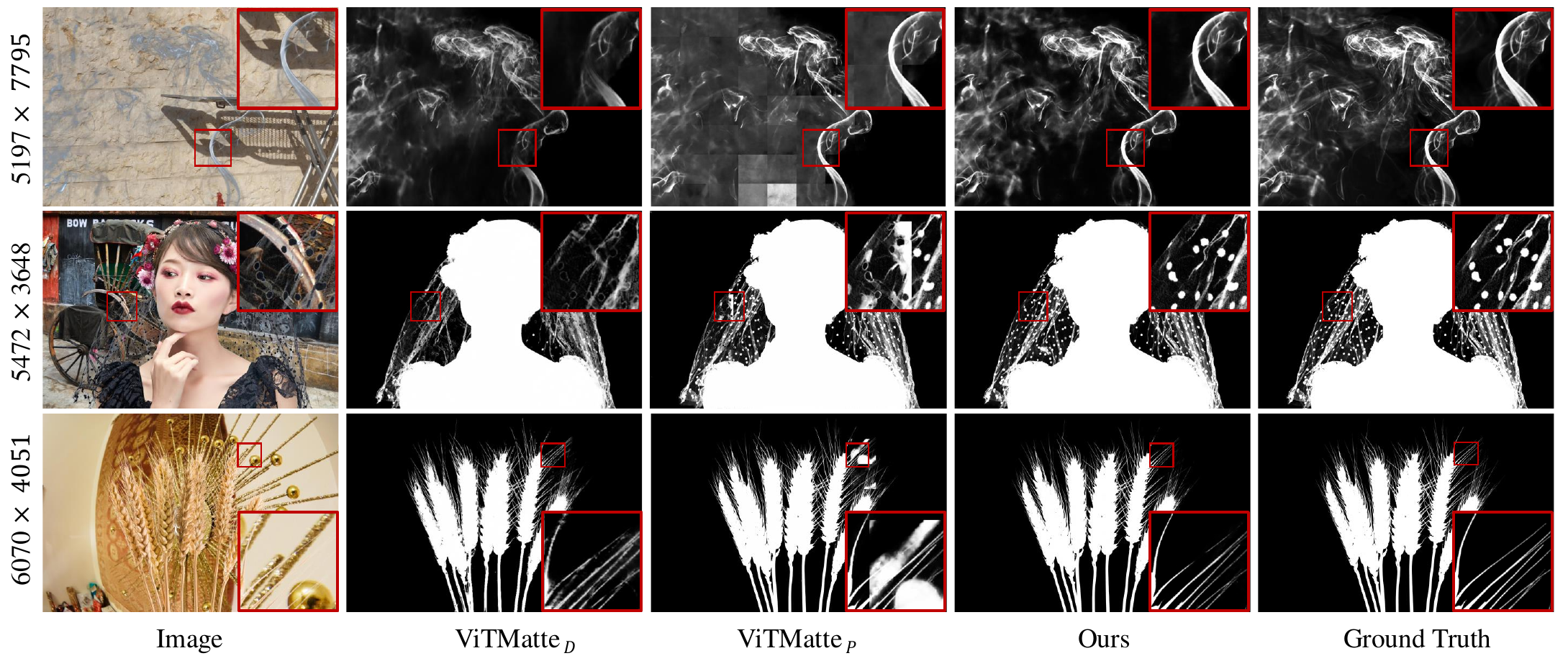}
    \caption{Qualitative comparison of the results on the UHR-395 test set. $D$ denotes downsampling the input and $P$ indicates dividing the input into patches. The resolution of each image is on the left.}
    \label{fig:uhrim}
\end{figure*}

%% file: Tables/UHRIM.tex
\begin{table}
\centering
\setlength{\tabcolsep}{5pt}
\begin{tabular}{l|cccc}
\toprule
Methods  & SAD $\downarrow$ & MSE $\downarrow$  &Grad $\downarrow$  & Conn $\downarrow$ \\ 
\midrule
FBA$_D$      & 1062.09   &  32.49  &   1196.34  & 1072.20   \\
FBA$_P$      & 1273.63   &  30.91  &   456.86  & 1296.05   \\
Matteformer$_D$      &  860.71  &  23.27  &  1158.63   & 823.68    \\
Matteformer$_P$ & 3386.12 & 22.83 & 2566.05 & 3478.81 \\
SparseMat      & 2779.50   & 20.92    &2079.01 &2324.84   \\
\midrule
ViTMatte-S$_D$      & 720.21   & 19.99     & 1026.79 & 668.02   \\
ViTMatte-S$_P$      & 843.87   & 17.06     &247.51 &820.58   \\
ViTMatte-B$_D$      & 669.44      & 17.72  & 983.17 & 622.50   \\
ViTMatte-B$_P$      & 727.36       & 12.79  &224.95 &693.02   \\
\midrule
\rowcolor[HTML]{d7dbdd} 
MEMatte-S      & 623.93   & 11.13     &220.09 &572.94   \\
\rowcolor[HTML]{d7dbdd} 
MEMatte-S$^*$      & \textbf{505.88}      & \textbf{8.50}  &\textbf{172.99} &\textbf{423.76}   \\

\bottomrule
\end{tabular}
\caption{Quantitative results on our proposed UHR-395. $D$ denotes downsampling the input and $P$ represents dividing the input into patches. $*$ indicates that the model is fine-tuned on our UHR-395 training set.}
\label{tab:uhrim}
\end{table}

%% file: Tables/PPT344.tex
\begin{table}
\centering
\begin{tabular}{l|cccc}
\toprule
     Methods  & SAD $\downarrow$ & MSE $\downarrow$  &Grad $\downarrow$ & Conn $\downarrow$ \\
\midrule
                        SparseMat      & 799.12   & 20.87 & 430.96 & 794.15    \\
                         FBA      & 202.94   & 20.74   & 80.14  & 213.16  \\
                         Matteformer      & 145.65   & 14.72   & 38.759 & 145.86    \\
                         
                        \midrule
                        
                         ViTMatte-S$_D$ & 166.33   & 13.46    &127.15 &159.40  \\
                         ViTMatte-S$_P$ & 141.98   & 12.80    &42.03 &142.31  \\
                         \rowcolor[HTML]{d7dbdd} 
                         MEMatte-S      & \textbf{117.05}   & \textbf{7.91}      &\textbf{33.38} & \textbf{113.03}  \\
                        \midrule
                         ViTMatte-B$_D$ & 155.14   & 12.51    &118.77 &147.82   \\
                         ViTMatte-B$_P$ & 151.86   & 15.02    &37.28 &152.47   \\
                         \rowcolor[HTML]{d7dbdd} 
                         MEMatte-B      & \textbf{114.95}   & \textbf{7.92}     &\textbf{29.85} &\textbf{109.54}   \\
                        \bottomrule
\end{tabular}
\caption{Quantitative results on PPT-344. $D$ denotes downsampling the input and $P$ indicates dividing the input into patches. Our method is marked in gray.}
\label{tab:PPT344}
\end{table}

%% file: Tables/aim500.tex
\begin{table}
\centering
\begin{tabular}{l|cccc}
\toprule
Methods  & SAD $\downarrow$ & MSE $\downarrow$ & Grad $\downarrow$ & Conn $\downarrow$ \\ \midrule
GCA      & 34.94   & 38.84   & 25.76    & 35.31    \\ 
MGMatting   & 55.14   &  134.35  &   40.55  &   53.67  \\ 
FBA      &  19.10  &  16.24  &   11.46  & 18.36    \\ 
Matteformer    & 26.90   & 29.01   & 23.00    & 26.63    \\ \hline
ViTMatte-S      & 19.57   & 15.8   & 12.78    & 18.85    \\ 
\rowcolor[HTML]{d7dbdd} 
MEMatte-S      & \textbf{18.88}   & \textbf{15.2}  & \textbf{12.42}    & \textbf{18.08}    \\\midrule
ViTMatte-B     & 17.33   & 16.48   & 14.17    & 16.30    \\ 
\rowcolor[HTML]{d7dbdd} 
MEMatte-B     & \textbf{16.99}   & \textbf{15.4}   & \textbf{13.93}    & \textbf{16.04}    \\ \bottomrule
\end{tabular}
\caption{Quantitative results on Real-world Dataset AIM500.}
\label{tab:aim500}
\end{table}

%% file: Figures/sparsity.tex
\begin{figure*}
    \centering
    \includegraphics[width=0.98\textwidth]{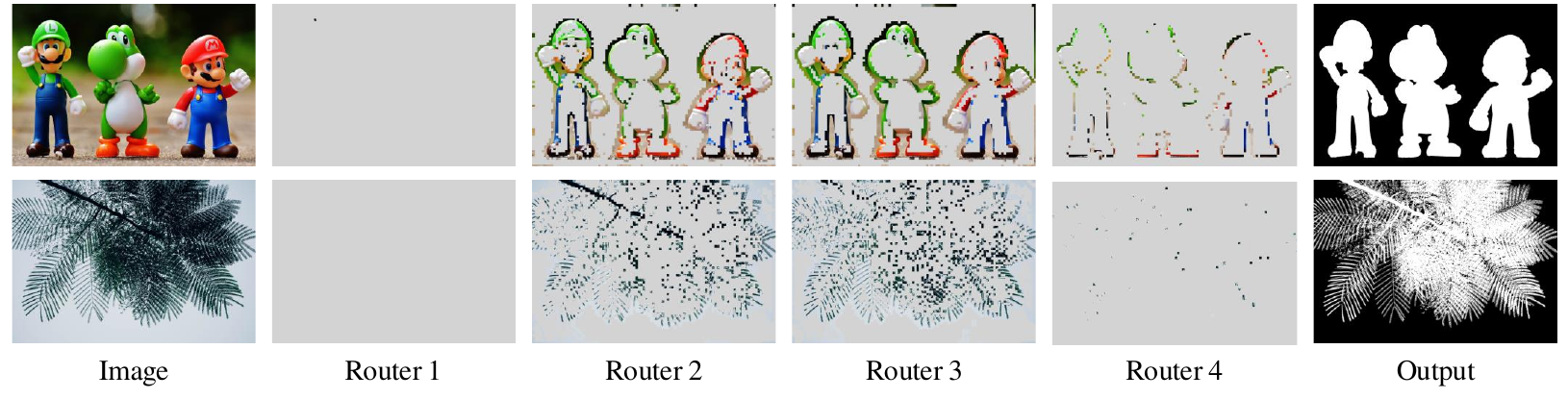}
    \caption{Visualization of the token routing. The retained tokens are routed to global attention branch, while the gray tokens are routed to the LTRM branch.  More visualization results are shown in supplementary materials.}
    \label{fig:sparsity}
\end{figure*}

%% file: Tables/routing_mechanism.tex
\begin{table}
\centering
\setlength{\tabcolsep}{4pt}
\begin{tabular}{l|cccc}
\toprule
Method      & SAD $\downarrow$ & MSE $\downarrow$ & Mem$_{/\text{GB}}$ & Latency$_{/\text{ms}}$  \\ \midrule

ViTMatte  & 21.46 & 3.30  & 6.20 & 186.00 \\  
\midrule
+ DynamicViT  & 33.47 & 9.16  & 4.20  & 104.35  \\
+ EViT        & 42.66 & 14.25  & 4.19  & 67.58  \\
+ ToMe        & 34.19 & 7.94  & 4.21  & 213.80  \\
\rowcolor[HTML]{d7dbdd}
+ BATR & \textbf{21.90} & \textbf{3.37}  & \textbf{0.71} & 84.99   \\
\bottomrule

\end{tabular}
\caption{Comparisons of different token compression mechanisms. Notably, all methods are based on the ViT-S backbone and are fine-tuned on the DIM training dataset.
}
\label{tab:routing_mechanism}
\end{table}

%% file: Figures/reduce_token_number_bar.tex
\begin{figure}[bth]
    \centering
    \captionsetup{aboveskip=10pt, belowskip=2pt}
    \includegraphics[width=0.47\textwidth]{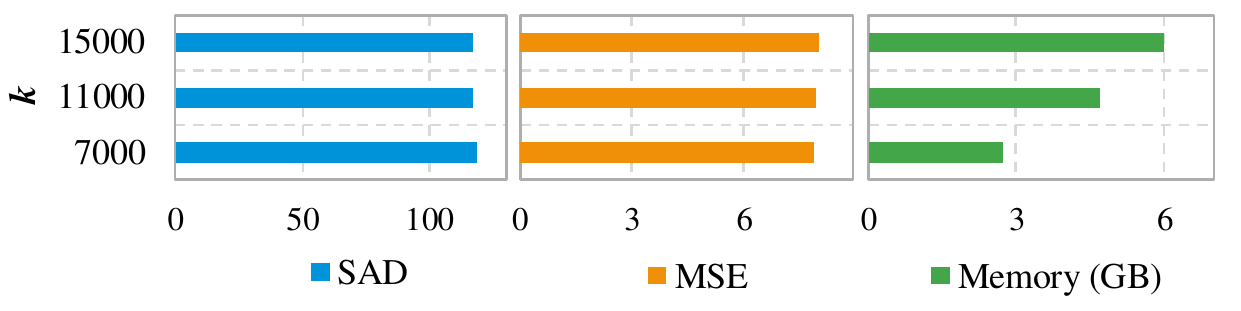}
    \caption{Effect of reducing maximum token number $k$.}
    \label{fig:reduce_token_number_bar}
\end{figure}

%% file: Tables/target_compression_degree.tex
\begin{table}[tb]
\centering
\begin{tabular}{c|cccc}
\toprule
$\rho$      & SAD $\downarrow$ & MSE $\downarrow$ & Mem$_{/\text{GB}}$ & Latency$_{/\text{ms}}$  \\ \midrule

0.75  & 21.78 & 3.32  & 4.22 & 166.09 \\
0.50  & 21.80 & 3.34  & 1.41  & 99.24  \\
\rowcolor[HTML]{d7dbdd}
0.25        & 21.90 & 3.37  & 0.71  & 84.99  \\
0.10        & 22.68 & 3.65  & 0.65  & 83.43  \\
\bottomrule

\end{tabular}
\caption{Comparisons of different target compression degree $\rho$ for training. The default setting of our method is marked in gray.
}
\label{tab:target_compression_degree}
\end{table}

%% file: Tables/components.tex
\begin{table}
\centering
\setlength{\tabcolsep}{5pt}
\begin{tabular}{cc|cccc}
\toprule
 Distill & LTRM & SAD $\downarrow$ & MSE $\downarrow$ & Mem$_{/\text{GB}}$ & Latency$_{/\text{ms}}$  \\ \midrule
           &  & 22.67 & 3.52  & 0.74    & 82.80 \\
  \ding{51} &  & 22.66 & 3.60  & 0.71   & 79.00   \\
            & \ding{51} & 22.54 & 3.55  & 0.75   & 83.17   \\
\rowcolor[HTML]{d7dbdd}
  \ding{51} & \ding{51} & \textbf{21.90} & \textbf{3.37}  & \textbf{0.71}   & 83.43   \\

\bottomrule

\end{tabular}
\caption{Effect of distillation loss and LTRM. Distill denotes distillation loss.
}
\label{tab:components}
\end{table}